\documentclass[twocolumn]{IEEEtran}

\usepackage[pdftex]{graphicx} 
\usepackage{amsmath} 
\usepackage{algorithmic}
\usepackage{array} 
\usepackage[caption=false,font=footnotesize]{subfig} 
\usepackage{dblfloatfix}
\usepackage{url} 

\usepackage{microtype}
\usepackage[utf8]{inputenc} 
\usepackage[T1]{fontenc}    
\usepackage{paralist} 
\usepackage{marvosym} 
\usepackage{bm} 
\usepackage{bbm} 
\usepackage{amssymb}   
\usepackage{mathrsfs} 
\usepackage{tabularray} 
\UseTblrLibrary{booktabs}       
\usepackage{tabularx} 
\usepackage[flushleft]{threeparttable} 
\usepackage{colortbl} 
\usepackage{makecell} 
\usepackage{multirow} 
\usepackage{nicefrac} 
\usepackage[dvipsnames]{xcolor} 
\usepackage{soul} 
\setul{0.5ex}{0.3ex}

\definecolor{citecolor}{RGB}{34,139,34}
\usepackage[pagebackref=true,breaklinks=true,colorlinks,
    citecolor=citecolor,bookmarks=false]{hyperref}
\usepackage{cleveref}  
\crefformat{figure}{Fig.~#2#1#3}
\crefformat{equation}{Eq.~(#2#1#3)}
\crefformat{table}{Table~#2#1#3}
\crefformat{section}{Section~#2#1#3}
\crefformat{algorithm}{Algorithm~#2#1#3}
\crefformat{appendix}{Appendix #2#1#3}
\crefformat{subsection}{subsection~#2#1#3}
\usepackage{cellspace}
\usepackage[numbers,sort&compress]{natbib} 

\usepackage{float}
\usepackage{xspace} 

\usepackage{pifont} 
\definecolor{Gray}{rgb}{0.9,0.9,0.9}
\definecolor{LightCyan}{rgb}{0.88,1,1}
\newcolumntype{a}{>{\columncolor{Gray}}c}

\begin{document}

\setlength{\abovedisplayskip}{.5\baselineskip} 
\setlength{\belowdisplayskip}{.5\baselineskip} 

\title{Zoom Out and Zoom In: Re-Parameterized Sparse Local Contrast Networks for Airborne \\ Small Target Detection}



\title{LAM-YOLO: Drones-based Small Object Detection on Lighting-Occlusion Attention Mechanism YOLO}

\author{
  Yuchen~Zheng,
  Yuxin~Jing,
  Jufeng~Zhao,
  Guangmang~Cui,

\thanks{Yuchen Zheng, YuXin Jing, Jufeng~Zhao and Guangmang~Cui are with the School of Electronics and Information, Hangzhou Dianzi University, Hangzhou, 310018, PR China (e-mail:232040254@hdu.edu.cn; jingyuxin22@163.com; dabaozjf@hdu.edu.cn; cuigm@hdu.edu.cn)(Corresponding author: Jufeng Zhao.).}

\thanks{This research was supported by Zhejiang Provincial Natural Science Foundation of China under Grant No. LY22F050002. This research was also supported by the Graduate Scientific Research Foundation of Hangzhou Dianzi University under Grant No. CXJJ2024070.}

}

\maketitle


\begin{abstract}

Drone-based target detection presents inherent challenges, such as the high density and overlap of targets in drone-based images, as well as the blurriness of targets under varying lighting conditions, which complicates identification. Traditional methods often struggle to recognize numerous densely packed small targets under complex background. To address these challenges, we propose LAM-YOLO, an object detection model specifically designed for drone-based. First, we introduce a light-occlusion attention mechanism to enhance the visibility of small targets under different lighting conditions. Meanwhile, we incroporate incorporate Involution modules to improve interaction among feature layers. Second, we utilize an improved SIB-IoU as the regression loss function to accelerate model convergence and enhance localization accuracy. Finally, we implement a novel detection strategy that introduces two auxiliary detection heads for identifying smaller-scale targets.Our quantitative results demonstrate that LAM-YOLO outperforms methods such as Faster R-CNN, YOLOv9, and YOLOv10 in terms of mAP@0.5 and mAP@0.5:0.95 on the VisDrone2019 public dataset. Compared to the original YOLOv8, the average precision increases by 7.1\%. Additionally, the proposed SIB-IoU loss function shows improved faster convergence speed during training and improved average precision over the traditional loss function.


\end{abstract}

\vspace{-1\baselineskip}

\section{Introduction} \label{sec:introduction}
%
\IEEEPARstart{U}nmanned Aerial Vehicles (UAVs) provide exceptional flexibility in aerial surveying, allowing for takeoff and landing at virtually any location and enabling the rapid completion of tasks. Their operational costs are significantly lower than those of traditional aircraft and helicopters. Consequently, drone-based target detection monitors are increasingly pivotal in fields such as mining surveys \cite{cao2023optimization}, traffic monitoring \cite{sun2022rsod}, and agricultural assessments \cite{barbedo2019review}. The detection of small targets in complex scenes captured by UAVs is particularly important.



\begin{figure}
    \centering
    \includegraphics[width=1.0\linewidth]{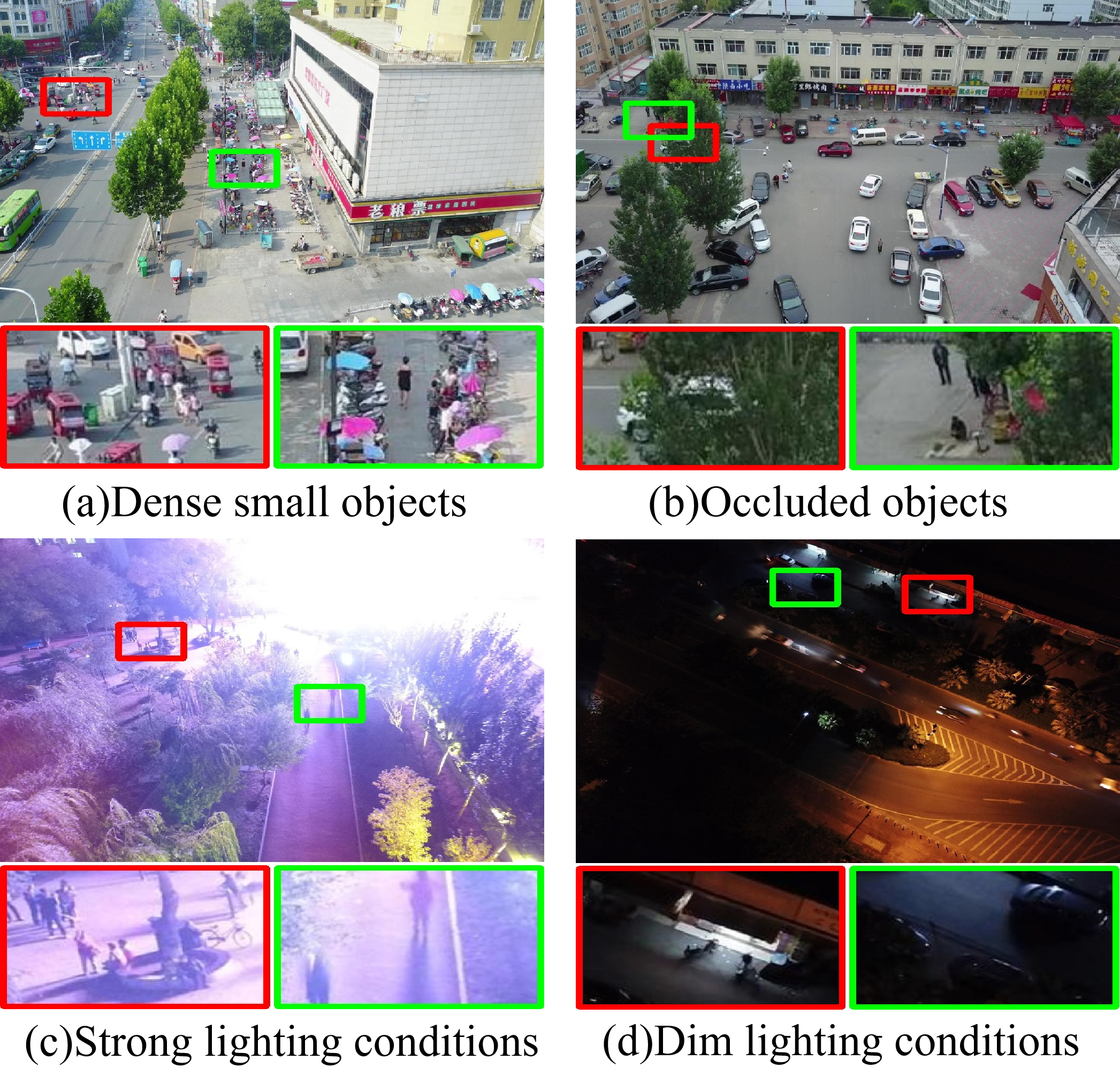}
    \caption{The complex background faced by  image target detection, the red and green bounding boxes are zoom in area: (a)Dense target of different types such as pedestrians and cars, (b)Target occlusion by other objects such as trees, (c)Strong lighting conditions of sunlight, and (d) Dim lighting conditions during nights.}
    \label{fig:fig1}
\end{figure}

However, the varying altitude and angles of drone capture image, combined with environmental factors, cause targets to exhibit different sizes and lighting conditions across images. Additionally, when the imaging distance is excessively large, small targets are susceptible to clustering, overlapping, and occlusion, which can result in missed detections and false positives\cite{dai2024background}. As illustrated in Fig. \ref{fig:fig1}(a)-(d), drone-based target detection faces significant challenges. Overall, the detection of small targets in drone imagery presents several notable obstacles:

\begin{enumerate}

\item{\textbf{Diverse Target Sizes:}} Drone-based target detection images often include targets of varying dimensions, with small objects particularly susceptible to interference. This complicates accurate identification and reduces detection accuracy.

\item{\textbf{Occlusion and Overlap:}} Small targets frequently experience occlusion and overlap, leading to missed detections and false alarms.

\item{\textbf{Environmental Variability:}} Variations in flight position, lighting conditions, weather, and sensor noise contribute to reduced contrast and sharpness of small targets in images, further complicating detection efforts.

\end{enumerate}

With advancements in deep learning techniques, convolutional neural networks (CNN) have emerged as prominent methods for target detection \cite{hu2024gradient, xu2021research}, particularly the YOLO (You Only Look Once) series methods \cite{chen2024det, wang2023uav}. To address the challenges mentioned above for drone-based target detection, several researchers have made concerted efforts in this area. Xu \emph{et al.} \cite{xu2024airborne}  employ spatial and channel attention prior to feature extraction to retain as much target information as possible. Chen \emph{et al.} \cite{chen2024det} address the low detection rates and high miss rates by enhancing the convergence speed without increasing the model parameters, utilizing the WIoU-v2 loss function for optimization. Although Wang \emph{et al.} \cite{wang2023uav} introduce a small target detection head to capture extremely small targets. However, these methods fail to consider that the addition of attention modules before the backbone network introduces more interference signals in the retained information, and whether the loss function can accurately regress between the ground truth and predicted bounding boxes ultimately affects the detection of extremely small targets.


Currently, the detection of small targets in drone-based aerial images has seen significant progress. While there are still unresolved issues in feature extraction that hinder high-precision on drone-based object detection, current approaches tend to serve as remedies for specific problems rather than fundamentally addressing the core challenges. The shortcomings of existing methods are shown in several key areas:1) the problem of missed and false detections is exacerbated. In drone-based aerial images, the scale of small targets varies significantly due to factors such as occlusion and distance. Without a specialized auxiliary detection head to enhance the recognition ability of these small targets, the model may miss detections. 2) Attention linear squeeze computing: Traditional attention mechanisms, utilizing linear computing similar to pooling in Squeeze and excitation networks (SENet) \cite{hu2018squeeze} and Pyramid Vision Transformer (PVT) \cite{wang2021pyramid}, unintentionally merge target features with primary background noise, diluting the target features within the background. To address these problems, we need to propose a learning paradigm that not only tackles the challenge of small targets changing size and scale with varying drone heights but also mitigates missed detections caused by lighting and occlusion affecting small targets in complex scenes through a specially designed attention mechanism.

To address the complexities faced by small targets in drone aerial imagery, this study aims to improve detection accuracy through enhancements in feature extraction, detection, and training strategies. We proposes LAM-YOLO, which builds upon YOLOv8 architecture by integrating multiple Lighting-Occlusion Attention Modules (LAM). This design facilitates the cross-scale integration of deep and shallow image information, thereby enhancing the model's capability for precise target localization. We utilizes Involution blocks to improve the interaction between feature maps of different scales, thereby allowing for the capture of subtle target characteristics. To optimize model training, we introduces an improved loss function that enhances the regression capability of predicted bounding boxes. Additionally, to address the uncertainty regarding target size, we incorporates an auxiliary detection head for small targets on top of the original detection head, thereby increasing the accuracy of detecting extremely small objects.


The main contributions of this article are listed as follows:

\begin{enumerate}

\item{\textbf{Enhanced Attention Mechanism}: We proposes the Lighting-Occlusion Attention Module (LAM), which includes channel attention, self-attention, and overlapping cross-attention mechanisms. Additionally, this study introduces these modules to strengthen multi-scale feature interaction, enabling the model to capture more information about small targets while also paying closer attention to occluded targets.}


\item{\textbf{Refined Regression Loss}: We integrate the SIB-IoU (Soft Intersection Bounding Box IoU) into the bounding box regression loss and utilizes scaling factors to generate auxiliary bounding boxes of varying sizes for loss computation. This approach facilitates faster and more efficient model training, thereby accelerating model convergence and improving precision}


\item{\textbf{Auxiliary Feature Detection}: To address the significant size variations of targets in drone-based images, we has designed an auxiliary feature detection strategy. Building upon the three existing detection heads in standard YOLOv8, this study adds two auxiliary detection heads specifically for extremely small targets, enhancing sensitivity and the ability to detect small objects.}


\item{\textbf{Experimental Validation}: We conducted detection experiments LAM-YOLO model. Results indicate a 7.1 \% improvement in accuracy on the VisDrone2019 dataset compared to baseline YOLOv8. Meanwhile, compare experiments also demonstrating superior detection performance to other state-of-the-art methods.}


\end{enumerate}


The remainder of this paper is organized as follows: Sec. \ref{sec:related} reviews several deep learning-based object detection methods. Sec. \ref{sec:method} provides a detailed description of the proposed model, while Sec. \ref{sec:experiment} presents the experimental results. Finally, Sec. \ref{sec:conclusion} concludes the paper.


\section{Related Work} 
\label{sec:related}

\subsection{Object Detection Methods}

\subsubsection{\textbf{Two Stage Detection Methods}}
Two-stage detection methods, such as the Faster R-CNN series \cite{ren2016faster}, initially propose explicit region proposals and subsequently classify these proposals for detection. Additionally, R2-CNN \cite{jiang2017r2cnn} incorporates a global context attention module and a proposal generator specifically designed for small targets, thereby enhancing the performance of the network.Many variants were also introduced later, such as MM R-CNN \cite{MM-RCNN}, MSA R-CNN \cite{MSA2024R-CNN}. While these methods achieve high detection accuracy, they suffer from slow detection speeds, rendering them inadequate for tasks with stringent real-time requirements.


\subsubsection{\textbf{One Stage Detection Methods}}
One-stage methods, such as CenterNet \cite{duan2019centernet}, CSPPartial-YOLO \cite{CSPPartial-YOLO}, EC-YOLOX \cite{EC-YOLOX}, LAR-YOLOv8 \cite{LAR-YOLOv8} and YOLO \cite{wang2023yolov7}, bypass the region proposal step and directly predict classification scores and bounding box regression offsets, aiming to achieve real-time performance while maintaining high accuracy. For example, the Transformer Prediction Head (TPH) \cite{zhu2021tph} adds an attention mechanism to the original YOLO, demonstrating great potential by assigning different weights to various regions of the feature map.


\subsubsection{\textbf{Transformer-Based Detection Methods}}
Transformer-based approaches utilize an attention model to establish dependencies among sequence elements, thereby enabling the extraction of global contextual information \cite{zhu2024towards}. Examples include the Vision Transformer (ViT) \cite{dosovitskiy2020image} and Swin Transformers \cite{liu2021swin}. As the first end-to-end target detection algorithm based on Transformers, DETR \cite{carion2020end} matches proposals with the ground truth one-to-one after feature extraction. Many variants were also introduced later, such as ARS-DETR \cite{ARS-DETR}, CDN-DETR \cite{CDN-DETR}, Deformable DETR \cite{Deformable2024DETR}. Despite the promising detection performance of Transformers, their complexity and substantial parameter count lead to significant consumption of computational resources and reduced inference efficiency.


Given the need for robust real-time performance in applications such as drone-based warning systems, we propose a drone image object detection algorithm from three perspectives to balance detection accuracy and inference efficiency: First, while striving for enhanced detection accuracy, it is essential to ensure real-time detection; therefore, the one-stage detection method is employed to extract spatial features of the targets. Second, we propose an efficient attention mechanism that emphasizes the features of small targets and utilizes multi-scale contextual information, thereby addressing the challenges of missed detection and occlusion of small targets. Finally, the loss function must be modified to better align predicted bounding boxes with the ground truth.

\subsection{YOLO Series Drone-based Object Detection}
The YOLO series methods \cite{redmon2016you, redmon2018yolov3, bochkovskiy2020yolov4, li2022yolov6, wang2023yolov7} are representative of single-stage object detection. Despite the considerable success of YOLO series models in object detection, specialized designs are still needed for complex scenarios in drone applications to improve detection accuracy. To address this, Lou \emph{et al.} \cite{lou2023dc} proposed a new downsampling method and modified the feature fusion network of YOLOv8 to enhance its detection capability for dense small targets. Wang \emph{et al.} \cite{wang2023uav} proposed the BiFormer attention mechanism, integrated into the backbone network, resulting in the Focus Faster Net Block (FFNB) feature processing module. This module introduces two new detection scales, significantly reducing the leakage rate. Xu \emph{et al.} \cite{xu2024airborne} proposed the YOLO-SA network, which integrates a YOLO model optimized for the downsampling step with an enhanced image set, achieving a new method for detecting small aerial targets by fusing RGB and infrared images. However, the existing detection accuracy of these methods still needs improvement for certain complex backgrounds, such as strong lighting or interference from dark night conditions, as well as for small objects like long-distance bicycles and pedestrians.


Current methods mainly focus on feature fusion in the detection head \cite{lou2023dc}, attention mechanisms in the bottleneck layer \cite{wang2023uav, lou2023dc}, and more powerful backbone feature extraction \cite{xu2024airborne}. However, no existing method specifically addresses the accurate detection of small targets at long distances in complex lighting and occlusion scenarios, which affects the robustness of drone-based target detection systems. To tackle this issue, we chose YOLOv8 as our foundational framework and integrated a specially designed lighting-occlusion attention mechanism (LAM). Additionally, we integrated an auxiliary small target detection head to enhance the model's capability for detecting extremely small targets. To enhance model training speed, we propose an improved loss function to regulate the generation of auxiliary bounding boxes of varying scales.


\subsection{Vision Attention Mechanisms}
The visual attention mechanism plays a crucial role in computer vision tasks \cite{zhu2021dau}, enabling high-quality feature representation by focusing on important elements while ignoring complex background interference and noise. The visual attention mechanism in deep learning is divided into channel and spatial attention mechanisms. The Squeeze-and-Excitation (SE) \cite{hu2018squeeze} block is a well-known module that dynamically focuses on channel characteristics. It uses global average pooling to compress channels into a single value, applies a non-linear transformation through a fully connected network, and multiplies the result by the input channel vector as a weight. ECA \cite{wang2020eca} reduces model redundancy and captures channel interactions by eliminating fully connected layers and using one-dimensional convolutional layers. Both SE and ECA apply attention mechanisms in the channel domain while ignoring the spatial domain. CBAM \cite{woo2018cbam} combines channel and spatial attention by using large kernel convolutions to aggregate positional information within a certain range. However, this design choice may lead to increased computational costs, making it less suitable for developing lightweight models. Additionally, a single layer with a large convolution kernel can capture only local positional information and cannot capture global positional information.

However, in drone-based target detection, we have designed a specialized attention mechanism to address occlusion and varying environmental lighting conditions. Specifically, we drew inspiration from research on the effects of environmental lighting on cognitive psychology \cite{liu2023image}. Our design is influenced by the regulatory role of light in human cognition, particularly its impact on alertness and attention. Consequently, our attention mechanism simulates this non-visual effect by dynamically adjusting the network's sensitivity to features under different lighting conditions, thereby enhancing the recognition of small targets in complex environments.




\section{Method} \label{sec:method}

\subsection{Overall Structure of LAM-YOLO}
\begin{figure}
\centering
\includegraphics[width=1\linewidth]{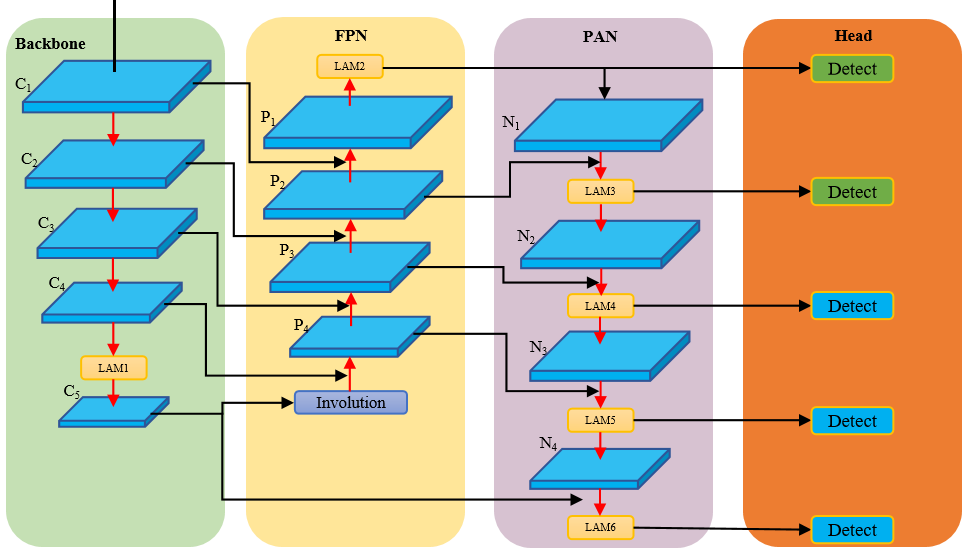}
    \caption{LAM-YOLO structure. The \textcolor{green}{green part} is the backbone, which contains CSPDarkNet\cite{wang2020cspnet}, the \textcolor{yellow}{yellow part} is FPN \cite{lin2017feature}, the \textcolor{purple}{purple part} is PAN \cite{PAN}, the \textcolor{orange}{orange part} is Head, the \textcolor{blue}{blue detection head} is the original YOLOv8, and the \textcolor{green}{green detection head} is the auxiliary detection head we introduced.}
    \label{fig:fig2}
\end{figure}

We propose an lighting-occlusion attention mechanism YOLO (LAM-YOLO) for complex environments drone-based small target detection, as shown in Fig. \ref{fig:fig2}. Building upon the YOLOv8, towards real-time detection, and faster training times while improving performance of higher precision. First, the Lighting-occlusion Attention Module (LAM) is introduced into the output layers of the backbone network and bottleneck. This module combines channel attention and self-attention mechanisms to enhance the lighting cognition ability of our module. Inspired by the Involution mechanism, we incorporate Involution \cite{li2021involution} blocks between the backbone and neck to enhance and share channel information, reducing information loss in the initial stages of the Feature Pyramid Network (FPN) \cite{lin2017feature}. Additionally, we proposed a novel soft inner-section bounding-box IoU (SIB-IoU) as the bounding box regression loss, which not only enhance performance and training time. Finally, we introduce two auxiliary detection heads with resolutions of 160×160 and 320×320 to improve the ability to detect small targets.

\begin{figure*}
    \centering
    \includegraphics[width=0.9\linewidth]{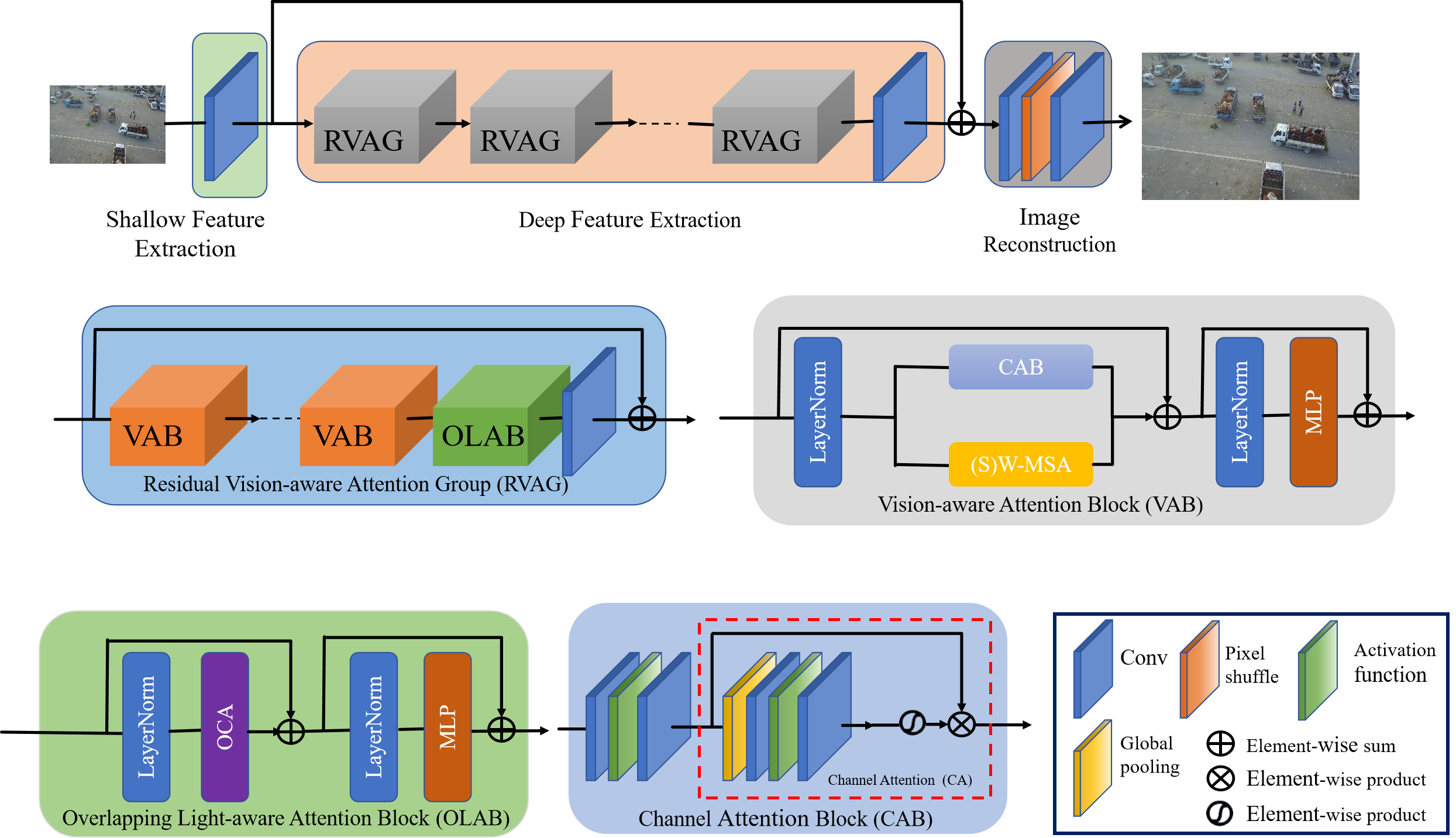}
    \caption{Lighting-occlusion attention module.The \textcolor{gray}{gray module} represents Residential Vision aware Attention Group (RVAG), the \textcolor{orange}{orange module} represents Vision aware Attention Block (VAB), the \textcolor{green}{green module} represents Overlapping Light aware Attention Block (OLAB)}
    \label{fig:fig3}
\end{figure*}

\subsection{Improvements In Network Structure}
\subsubsection{\textbf{Lighting-occlusion Attention Module}}
Due to the long range of the image taken by the drone-based, the proportion of small objects is high, and most of the small objects are easily to be occlusion. In order to better paid attention to by the detection model. We proposed to a hybrid attention module called Lighting-occlusion attention (LAM) in the backbone network of the model and the output part of neck. LAM combines channel attention and windows-based self-attention mechanisms, by fully take advantage of their complementary advantages, leveraging global statistical information and strong local feature extraction capabilities. 

The overall architecture of LAM consists of three parts: Shallow feature extraction, deep feature extraction, and image reconstruction. For a given low-resolution input $I_{LR} \in \mathbb{R}^{H \times W \times C_{in}}$, we first use a convolutional layer to extract shallow features $F_0 \in \mathbb{R}^{H \times W \times C}$, where $C_{in}$ and $C$ represent the number of input channels and intermediate features, respectively. Next, we use a series of Residual Hybrid Attention Groups (RHAG) and a $3 \times 3$ convolutional layer $H_{\text{conv}}(\cdot)$ for deep feature extraction. We add global residual connections to integrate the shallow features $F_0$ and the deep features $F_D \in \mathbb{R}^{H \times W \times C}$. Finally, the high-resolution results are reconstructed by the reconstruction module. As shown in Fig. \ref{fig:fig3}, each RHAG contains several mixed attention blocks (HAB), overlapping cross attention blocks (OCAB), and $3 \times 3$ convolutional layers with residual connections. For the reconstruction module, the pixel scrambling method\cite{shi2016real-time} is used to upsample the fused features.

\subsubsection{\textbf{Block of Involution}}
\begin{figure}
        \centering
        \includegraphics[width=1.0\linewidth]{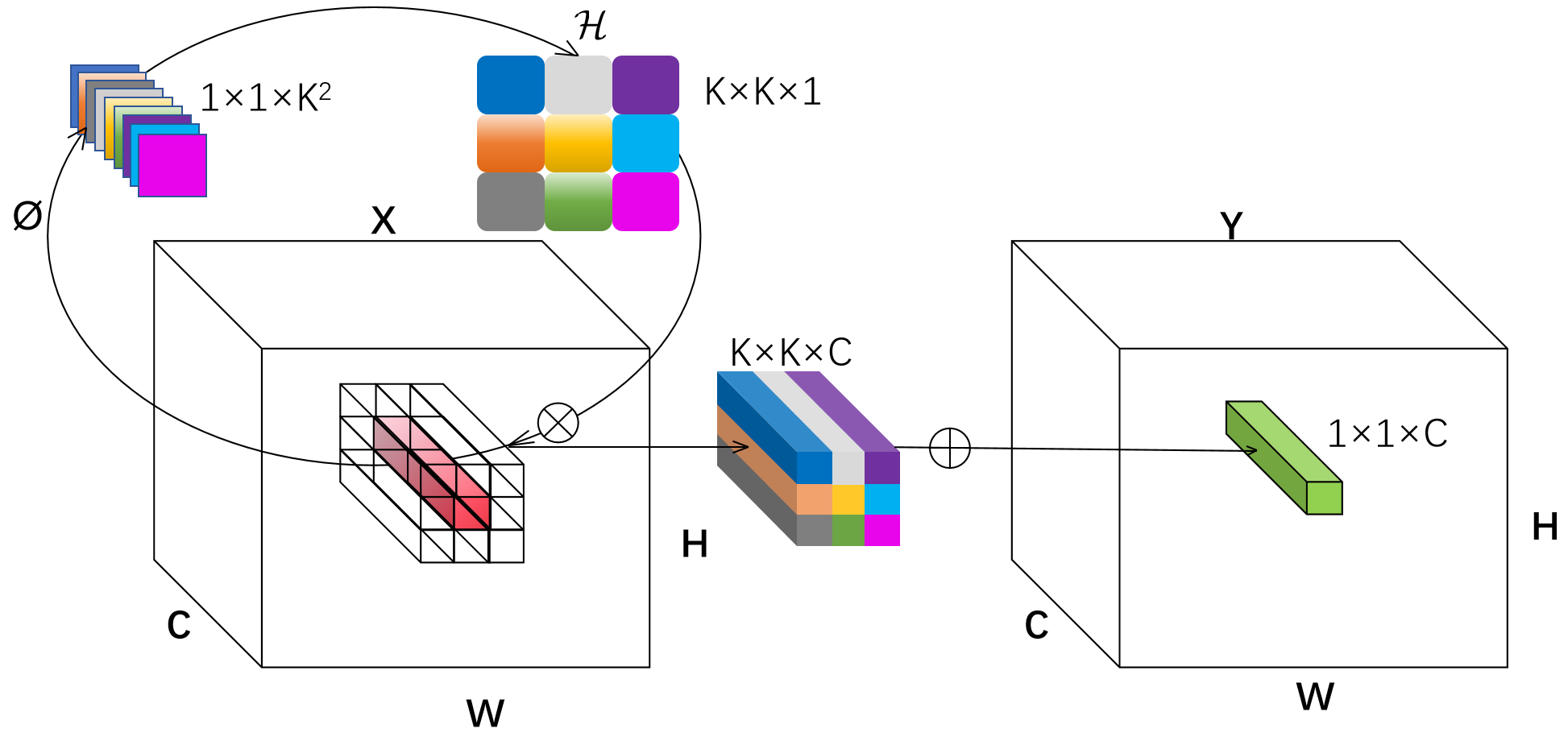}
        \caption{The structure of the Involution block}
        \label{fig:fig4}
\end{figure}

In the multi-scale feature fusion process, feature maps at different scales may not fully interact, resulting in the model unable to make full use of information at different scales to detect small objects \cite{lou2023dc}. To solve this problem, we add Involution block to Neck, which helps to generate richer feature representations by redistributing information across channel and spatial dimensions, thereby improve the ability of LAM-YOLO to understand the image content and perform well when dealing with small and dense objects. The Involution block is shown in Fig. \ref{fig:fig4}. Involution kernel $\mathcal{H} \in \mathbb{R}^{H \times W \times K \times K \times G}$ aim to combine transformations that exhibit inverse properties in spatial and channel domains, where $H$ and $W$ represent the height and width of the feature map, $K$ is the kernel size, and $G$ is the number of groups. Each group shares the same Involution kernel. Specifically, for the pixel $X_{ij} \in \mathbb{R}^{C}$ (Omit the subscript of $C$), a specific Involution kernel is designed, denoted as $h_{ij,:,:,:}^g \in \mathbb{R}^{K \times K}$, for $g = 1, 2, ..., G$, shared across all channels. Finally, the output feature map $Y_{ij,:}$ of the Involution is obtained by the following formula:
				\[
				\mathbf{Y}_{i,j,k} = \sum_{(u,v) \in \Delta_K} \mathcal{H}_{i,j,u+\lfloor K/2 \rfloor, v+\lfloor K/2 \rfloor, \lfloor kG/C \rfloor} \mathbf{X}_{i+u,j+v,k}
				\tag{1}
				\]
				\[
				\mathcal{H}_{i, j}=\phi\left(\mathbf{X}_{\Psi_{i, j}}\right)
				\tag{2}
				\]	
Where u and v are offset indices of the kernel in the two-dimensional space (height and width). $\mathbf{\Psi_{i, j}}$ indexes the set of pixels $\mathbf{{H}_{i, j}}$ is conditioned on.

In Fig. \ref{fig:fig4} to facilitate the demonstration of setting G to 1, $\otimes$denotes the broadcast multiplication across C channels, and $\oplus$denotes the sum operation aggregated in a $K \times K$spatial neighborhood. Thus, the information contained in the channel dimension of a single pixel is implicitly dispersed near its space, which is useful for obtaining rich receptive field information.

\subsubsection{\textbf{Soft Inner-section Bounding-box IoU}}
In drone-captured images, the proportion of small objects is quite high, and a well-designed loss function can enhance detection performance. YOLOv8 calculates the bounding box regression loss using Distribution Focal Loss(DFL) \cite{Li2020} and Complete-IoU(CIoU) \cite{Zheng2020}, but CIoU does not balance hard and easy samples when computing the loss. Additionally, CIoU uses the aspect ratio as one of the penalty factors; however, if the ground truth box and the predicted box share the same aspect ratio but have different width and height values, the penalty term cannot accurately reflect the actual difference between the two. These points hinder its ability to effectively identify small objects in complex aerial scenes captured by drones.
				
SCYLLA-IoU(SIoU) \cite{Gevorgyan2022} introduces the vector Angle between the real box and the predicted box for the first time, redefines the penalty index, improves the speed of training and the accuracy of inference, and accelerates the convergence speed of the model. Considering the influence of Angle between Anchor box and GT box on bounding box regression, the Angle loss is introduced into the bounding box regression loss function, and the related formula is as follows:
\begin{equation}
    L_{SIoU} = 1 - IoU + \frac{\Delta + \Omega}{2}
    \tag{3}
\end{equation}
Where $\Delta$is the distance loss, $\Omega$ is the shape loss.

Angle loss $\Lambda$ represents the minimum angle between the connection of the center point of the GT box:
\begin{equation}
    \Lambda = \sin \left( 2 \sin^{-1} \left( \frac{\min \left( \left| x_c^{gt} - x_c \right|, \left| y_c^{gt} - y_c \right| \right)}{\sqrt{ \left( x_c^{gt} - x_c \right)^2 + \left( y_c^{gt} - y_c \right)^2 + \epsilon }} \right) \right)
    \tag{4}
\end{equation}
Where $x_c^{gt}, y_c^{gt}$represents the coordinates of the center point $b^{gt}$ of object $B^{gt}$ in the ground truth, and $x_c, y_c$represent the coordinates of the center point $b$ of object $B$ in the prediction. $\epsilon$ represents a very small positive number to prevent the denominator from appearing as 0.

This term is intended to adjust the anchor box towards the nearest coordinate axis, prioritizing alignment with either the X-axis or Y-axis depending on the change in angle. When the angle is 45°, $\Lambda$ equals 1. If the center points are aligned with the X or Y axis, $\Lambda$ equals 0. After considering the Angle cost, the definition of distance loss is as follows:
\begin{equation}
    \Delta = \frac{1}{2} \sum_{t=w,h} \left( 1 - e^{-\gamma \rho_t} \right), \quad \gamma = 2 - \Lambda
    \tag{5}
\end{equation}

\begin{equation}
    \left\{
    \begin{aligned}
        \rho_x &= \left( \frac{b_x - b_x^{gt}}{w^c} \right)^2 \\
        \rho_y &= \left( \frac{b_y - b_y^{gt}}{h^c} \right)^2
    \end{aligned}
    \right.
    \tag{6}
\end{equation}
where $\rho_t$ represents the normalized distance of the bounding box width or height, and $\gamma$ is a coefficient adjusted based on the angle loss $\Lambda$. $\rho_x,\rho_y$ represent the normalized distances of the center point of the bounding box in the x and y directions, respectively. $b_x,b_y$ represent the coordinates of the predicted bounding box center point. $b_x^{gt},b_y^{gt}$ represents the coordinates of the center point of the real bounding box. $w^c,h^c$ represent the width and height of the minimum bounding box covering the target box and the predicted box.

The shape loss $\Omega$ mainly describes the size difference between GT box and Anchor box:
\[
\Omega = \frac{1}{2} \sum_{t=w,h} \left(1 - e^{\omega_t}\right)^\theta, \quad \theta = 4
\tag{7}
\]
\[
\begin{cases}
    \omega_w = \frac{|w - w_{gt}|}{\max(w, w_{gt})} \\
    \omega_h = \frac{|h - h_{gt}|}{\max(h, h_{gt})}
    \tag{8}
\end{cases}
\]
where $\omega_t$ represents the normalized difference in width or height of the bounding box. $\omega_w,\omega_h$
represent the normalized differences in bounding box width and height, respectively. $w,h$ represent the width and height of the predicted bounding box. $ w_{gt},h_{gt}$ represents the width and height of the real bounding box.

\begin{figure}
    \centering
    \includegraphics[width=1.0\linewidth]{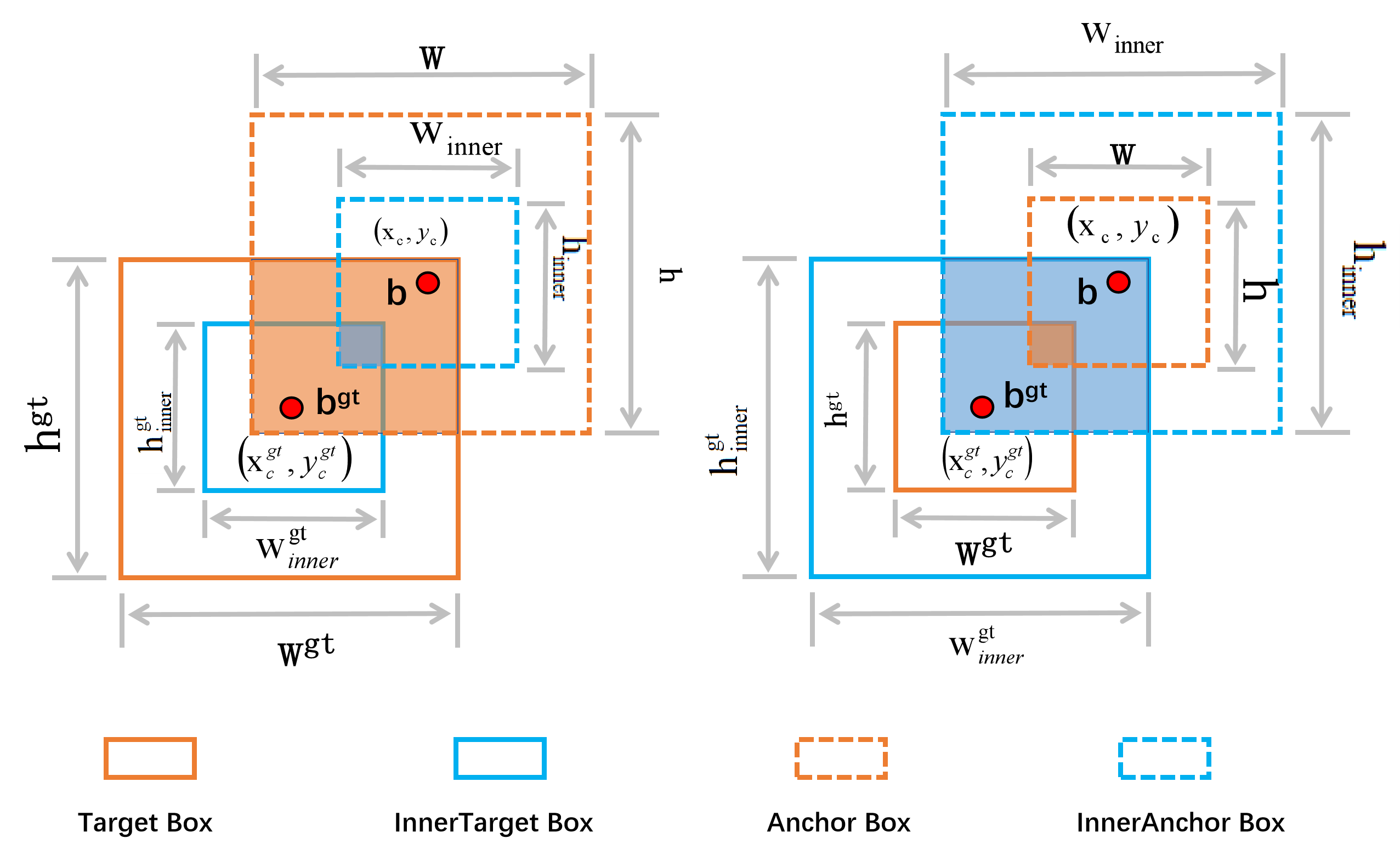}
    \caption{Inner-IoU description}
    \label{fig:fig5}
\end{figure}

SIB-IoU introduces the scale factor control to generate auxiliary bounding boxes of different scales to calculate the loss, and applies it to the existing IoU. As shown in Fig. \ref{fig:fig5}, the GT box and the anchor box are represented as $b^{gt}$ and $b$, with the center coordinates of the GT box denoted by $(x_c^{gt}, y_c^{gt})$ and those of the anchor box by $(x_c, y_c)$. The width and height of the GT box are represented by $w^{gt}$ and $h^{gt}$, respectively, while the width and height of the anchor box are denoted by $w$ and $h$. The scaling factor, denoted as the variable ratio, typically ranges from [0.5, 1.5]. When this ratio is less than 1, indicating that the auxiliary bounding box is smaller than the actual bounding box, the effective regression range is narrower compared to IoU loss. However, the gradient's absolute value is larger than that from IoU loss, which facilitates faster convergence for high IoU samples. Conversely, when the ratio exceeds 1, the larger-scale auxiliary bounding box extends the effective regression range and improves the regression performance for low IoU samples. The relevant formula is as follows:
\begin{align}
b_{l}^{g t}=x_{c}^{g t}-\frac{w^{g t} * \text { ratio }}{2}, b_{r}^{g t}=x_{c}^{g t}+\frac{w^{g t} * \text { ratio }}{2}\tag{9} \\
b_{t}^{g t}=y_{c}^{g t}-\frac{h^{g t} * \text { ratio }}{2}, b_{b}^{g t}=y_{c}^{g t}+\frac{h^{g t} * \text { ratio }}{2} \tag{10} 
\end{align}
\begin{align}
b_{l}=x_{c}-\frac{w * \text { ratio }}{2}, b_{r}=x_{c}^{g t}+\frac{w * \text { ratio }}{2}\tag{11} \\
b_{t}=y_{c}-\frac{h * \text { ratio }}{2}, b_{b}=y_{c}^{g t}+\frac{h * \text { ratio }}{2}\tag{12} 
\end{align}
\begin{equation}
    \small
    \text{inter} = (\min(b_r^{gt}, b_r) - \max(b_l^{gt}, b_l)) (\min(b_b^{gt}, b_b) - \\
    \max(b_t^{gt}, b_t))  
    \normalsize\tag{13}\\
\end{equation}
\begin{equation}
\text{union} = (w^{gt} \times h^{gt}) \times (\text{ratio})^2 + (w \times h) \times (\text{ratio})^2 - \text{inter}\tag{14} \\
\end{equation}
\begin{align}
IoU^{inner} &= \frac{inter}{union} \tag{15}
\end{align}
SIoU and Inner-IoU are combined to obtain the formula as follows:
\begin{align}
L_{\text {SIB-IoU }}=L_{S I o U}+I o U-I o U^{\text {$inner$}}\tag{16}
\end{align}	
where $b_{l}^{g t},b_{r}^{g t}$ represents the x-coordinate of the left and right edges of the real bounding box. $b_{t}^{g t},b_{b}^{g t}$ represents the y-coordinate of the upper and lower edges of the real bounding box. $x_{c}^{g t},y_{c}^{g t}$ represents the x and y coordinates of the center point of the real bounding box.  $w^{g t},h^{g t}$ represents the width and height of the real bounding box. ratio represents the scaling factor. $b_{l},b_{r}$ represents the x-coordinates of the left and right edges of the predicted bounding box. $b_{t},b_{b}$ represents the y-coordinate of the upper and lower edges of the predicted bounding box. $x_{c},y_{c}$ represents the x and y coordinates of the predicted bounding box center point. $w,h$ represents the width and height of the predicted bounding box. $inter$ epresents the intersection area between the auxiliary bounding box and the real bounding box. $union$ represents the union area of the auxiliary bounding box and the real bounding box. $IoU^{inner}$ represents the intersection to union ratio between the auxiliary bounding box and the real bounding box.

\AtBeginEnvironment{table}{\footnotesize} 

\section{Experiments and Analysis} \label{sec:experiment}
In Sec. \ref{sec:experiment}, we conducted experiments on a drone-based dataset and compared the results of state-of-the-art (SOTA) methods to evaluate the performance of LAM-YOLO. This study is complemented by a detailed ablation analysis to evaluate the contribution of each component in LAM-YOLO. We aim to address the following pivotal questions in our experimental analysis:
\begin{enumerate}

\item{\textbf{Q1:Comparative Analysis with State-of-the-Art}}: We compare LAM-YOLO with other cutting-edge methods, focusing on quantitative assessments and visual results in Sec. \ref{subsec:sota}.

\item{\textbf{Q2:Module-wise Performance Gain}}: Sec. \ref{subsec:ablation} provides a detailed account of the individual performance enhancements attributed to each LAM-YOLO's modules.

\item{\textbf{Q3:Optimal LAM Number}}: We compare LAM-YOLO with other cutting-edge methods, focusing on quantitative assessments and visual results in Sec. \ref{subsec:lamablation}.

\item{\textbf{Q4:The impact of the loss function}}:  In Sec. \ref{subsec:lossablation}, we validated the impact of improving the loss function on training, and statistically analyzed the relationship of evaluation metrics and training epochs.

\end{enumerate}

\subsection{Experimental Settings} \label{subsec:setting}

\subsubsection{\textbf{Dataset}}
\begin{figure}
    \centering
    \includegraphics[width=1.0\linewidth]{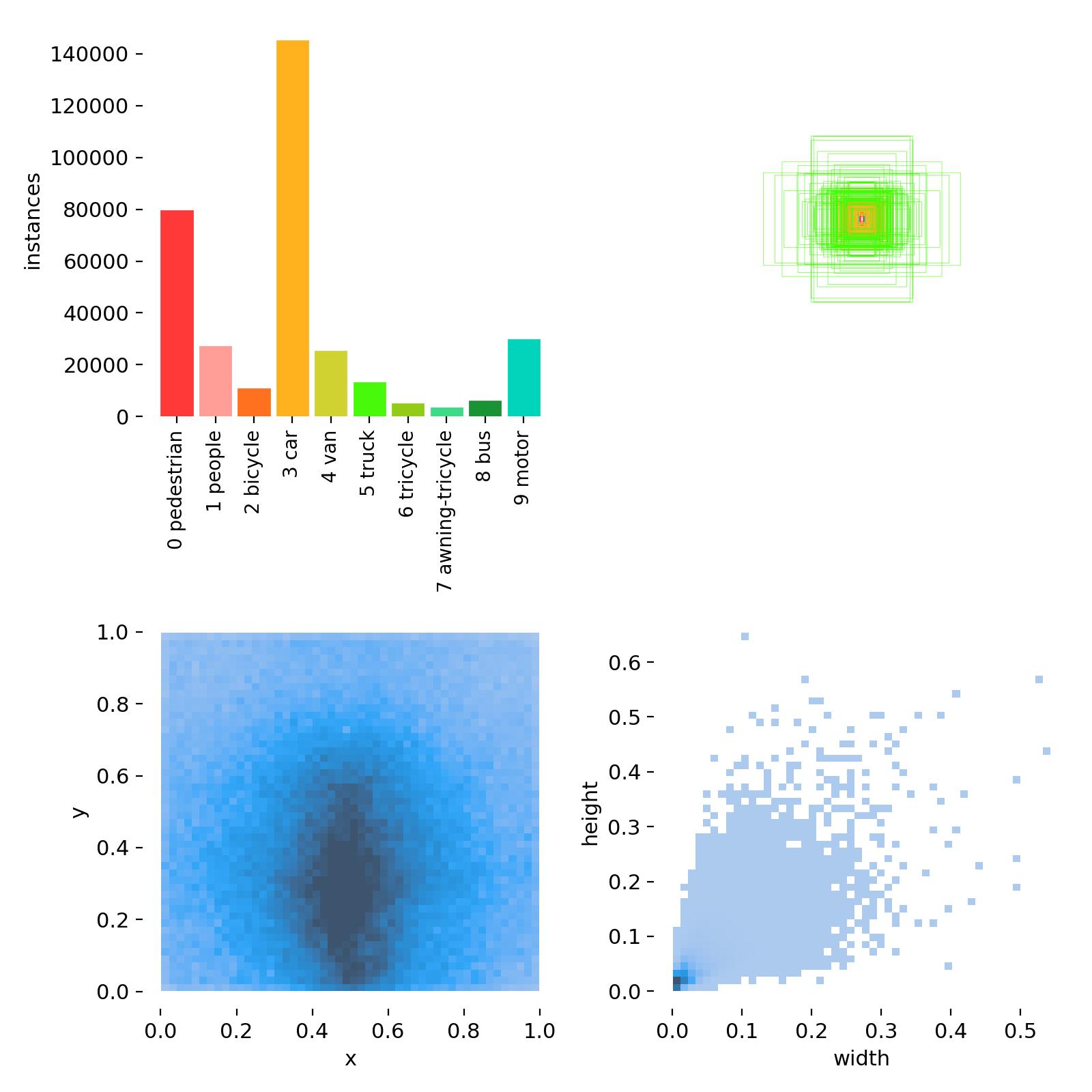}
    \caption{Information about the manual annotations of the objects in the VisDrone2019 dataset}
    \label{fig:fig7}
\end{figure}

We employ VisDrone2019 \cite{wang2023uav} public data sets, which captured by various professional drone cameras in 14 scenes encompass different locations (urban and rural) and densities (sparse and crowded). The VisDrone2019 dataset includes 10 types of detection targets, such as pedestrians, vehicles, and bicycles; however, the distribution of class samples is highly uneven, Fig. \ref{fig:fig7} illustrates the sample statistics of class distribution. In this figure, the vertical axis represents the height of the bounding box, while the horizontal axis denotes its width. Observations indicate that points are concentrated in the lower-left quadrant, suggesting the dominance of smaller objects within the VisDrone2019 dataset. This trend reflects practical applications of drone-based technologies and aligns well with the research context and issues addressed in this paper.

\subsubsection{\textbf{Implementation Details}}
In this experiment, a 20 vCPU Intel(R) Xeon(R) Platinum 8457C processor with L20 GPU and 48G running memory was used. Some of the key parameter settings during model training are shown in Tab. \ref{tab:Parameters}.
\begin{table}[h]
    \centering
    \footnotesize
    \caption{Some key parameters set during model training}
    \label{tab:Parameters}
    \renewcommand{\arraystretch}{1.2} 
    \begin{tabular}{ll}
        \hline
        Parameters & Setup \\ \hline
        Epochs & 300 \\
        Image size & 640 \\
        Batch size & 8 \\
        Weight decay & 0.0005 \\
        Momentum & 0.937 \\
        Optimizer & Adam \\
        Ratio (SIB-IoU) & 1.15 \\ \hline
    \end{tabular}
\end{table}
        
\subsubsection{\textbf{Evaluation Metrics}}
In order to evaluate the detection performance of the improved model, we focused on the accuracy and complexity of the model, and took accuracy, recall rate, $mAP0.5$, $mAP0.5:0.95$.
        
Precision is defined as the ratio of the number of positive samples predicted by the model to the total number of detected samples, and is calculated as follows:
\begin{equation}
    \text{Precision} = \frac{TP}{TP + FP}\tag{17}
\end{equation}
    
The recall rate is the proportion of correctly predicted positive samples to the total actual positive samples:
\begin{equation}
    \text{Recall} = \frac{TP}{TP + FN}\tag{18}
\end{equation}
where $TP$ (True Positive) denotes the count of accurately positive samples, specifically those predicted as positive and confirmed as positive. $FP$ (False Positive) indicates the count of negative samples inaccurately classified as positive, namely those that are actually negative yet predicted as positive. $TN$ (True Negative) signifies the count of accurately classified negative samples, specifically those predicted as negative and confirmed as negative. $FN$ (False Negative) denotes the count of positive samples inaccurately classified as negative, namely those that are actually positive yet predicted as negative.

The average precision (AP) is equal to the area under the precision–recall curve:

\begin{equation}
    AP = \int_{0}^{1} \text{Precision}(\text{Recall}) \, d(\text{Recall})\tag{19}
\end{equation}

The mean precision (mAP) is the average of the $AP$ values for all sample categories, indicating the effectiveness of the trained model in detecting all categories:
\begin{equation}
    \text{mAP} = \frac{1}{N} \sum_{i=1}^{N} AP_i\tag{20}
\end{equation}
Where $AP_i$ represents the $AP$ value of the i-th category, and $N$ represents the number of categories in the VisDron2019 training dataset ($N=10$).

\begin{table}[ht]
  \renewcommand\arraystretch{1.2}
  \footnotesize
  \centering
  \caption{Comparison with Other State-of-the-art methods on four datasets. The $\uparrow$ indicates that the higher the indicator, the better.We display the best result in the \textcolor{red}{red} color and the second-best result in the \textcolor{blue}{blue} color, the third-best result in \textcolor{green}{green} color.}
  \label{tab:models_comparison}
  \setlength{\tabcolsep}{3.5pt}
  \begin{tabular}{l|c|ccc}
    Method & Backbone & mAP0.5/\% $\uparrow$ & mAP0.5:0.95/\%$\uparrow$ \\
  \Xhline{1pt}
  \multicolumn{4}{l}{\textit{One-stage Detection Method}}  \\
  \hline
  TOOD \cite{TOOD} & ResNet50 & 37.7 &23.2\\
  YOLOX-Tiny \cite{YOLOX}& ResNet50 & 29.6 &16.0\\
  CenterNet \cite{Centernet}& ResNet50 & 34.9 &21.0\\
  YOLOv9 \cite{yolov9}& CSPDarknet & 47.7 &29.3\\
  YOLOv10 \cite{yolov10}& CSPDarknet & 38.5 &22.9\\
  RetinaNet \cite{RetinaNet}& ResNet50 & 32.0 &19.3\\
  GFL \cite{GFL}& ResNet50 & 36.2 &22.0\\
  Rtmdet \cite{Rtmdet}& CSPNet & 32.7 &19.2\\
  \hline  
  \multicolumn{4}{l}{\textit{Two-stage Detection Method}}  \\  
  \hline
  Faster R-CNN \cite{FasterR-CNN}& ResNet50 & 47.6 &29.2\\
  Cacade R-CNN \cite{CascadeR-CNN}& ResNet50 & \textcolor{blue}{48.1} &\textcolor{blue}{29.6}\\
  Grid R-CNN \cite{Grid-RCNN}& ResNet50 & \textcolor{green}{47.9} &\textcolor{green}{29.5}\\
  \hline
  \multicolumn{4}{l}{\textit{End-to-end Detection Method}}  \\
  \hline
  DETR \cite{DETR}& ResNet50 & 34.3 & 17.6\\
  DAB-DETR \cite{DAB-Detr}& ResNet50 & 41.3 & 25.4\\
  \hline
  \rowcolor[rgb]{0.9,0.9,0.9}$\star$ \textbf{LAM-YOLO(Ours)}  & CSPDarknet & \textcolor{red}{48.8} & \textcolor{red}{29.9} \\
\end{tabular}
\end{table}

\begin{table*}[htbp]
    \centering
    \footnotesize
    \clearpage
    \caption{Comparison of Models with Different Techniques.P2 represents a feature map size of 160 * 160, P1 represents a feature map size of 320 * 320, and LAMi (i=1, 2,... 6) represents LAMs at different positions. $\uparrow$ indicates that the higher the indicator, the better}
    \label{tab:Ablation experiment}
    \resizebox{\linewidth}{!}{ 
        \begin{tabular}{>{\bfseries}ccccccccccccc}
            \hline
             \textbf{P2(160*160)} & \textbf{P1(320*320)} & \textbf{SIB-IoU} & \textbf{Involution} & \textbf{LAM1} & \textbf{LAM2} & \textbf{LAM3} & \textbf{LAM4} & \textbf{LAM5} & \textbf{LAM6} & \textbf{mAP0.5/\% $\uparrow$} & \textbf{mAP0.5:0.95/\% $\uparrow$} \\
            \hline
            \checkmark & & & & & & & & & &45.1 &27.0  \\
            \hline
            \checkmark &\checkmark & & & & & & & & &46.7 &27.9  \\
            \hline
            \checkmark &\checkmark &\checkmark & & & & & & & & 47.4 & 28.4  \\
            \hline
            \checkmark &\checkmark &\checkmark &\checkmark & & & & & & & 47.6 & 28.7  \\
            \hline
            \checkmark &\checkmark &\checkmark &\checkmark &\checkmark & & & & & & 47.7 & 28.9  \\
            \hline
            \checkmark &\checkmark &\checkmark &\checkmark &\checkmark & & & & &\checkmark  & 48.0 & 29.1  \\
            \hline
            \checkmark &\checkmark &\checkmark &\checkmark &\checkmark & & & &\checkmark &\checkmark  & 48.1 & 29.2  \\
            \hline
            \checkmark &\checkmark &\checkmark &\checkmark &\checkmark & & &\checkmark &\checkmark &\checkmark  & 48.1 & 29.4  \\
            \hline
            \checkmark &\checkmark &\checkmark &\checkmark &\checkmark & &\checkmark &\checkmark &\checkmark &\checkmark  & 48.2 & 29.5  \\
            \hline
            \checkmark &\checkmark &\checkmark & &\checkmark &\checkmark &\checkmark &\checkmark &\checkmark &\checkmark  & 48.7 & 29.5  \\
            \rowcolor[rgb]{0.9,0.9,0.9} \hline
            \checkmark &\checkmark & \checkmark & \checkmark & \checkmark & \checkmark & \checkmark & \checkmark & \checkmark & \checkmark & \textbf{48.8} & \textbf{29.9} \\
            \hline
        \end{tabular}
    }
\end{table*}

\subsection{Comparison with State-of-the-Arts} \label{subsec:sota}
\subsubsection{\textbf{Quantitative comparison and analysis}}
We compared our model with current advanced methods and state-of-the-art (SOTA) methods, including one-stage detection methods: TOOD \cite{TOOD}, YOLOX-Tiny \cite{YOLOX}, CenterNet \cite{Centernet}, YOLOv9 \cite{yolov9}, YOLOv10 \cite{yolov10}, RetinaNet \cite{RetinaNet}, GFL \cite{GFL}, and Rtmdet \cite{Rtmdet}; two-stage detection methods: Faster R-CNN \cite{FasterR-CNN}, Cascade R-CNN \cite{CascadeR-CNN}, and Grid R-CNN \cite{Grid-RCNN}; and end-to-end detection methods: DETR \cite{DETR}, DAB-DETR \cite{DAB-Detr}. Tab.  
   \ref{tab:models_comparison} presents the quantitative comparison results of multiple object detection models across various evaluation metrics. The evaluation metrics include mAP@0.5 and mAP@0.5-0.95, which respectively measure the average accuracy of the model at different thresholds. mAP@0.5 represents the average accuracy when IoU is 0.5, while mAP@0.5-0.95 reflects the average mAP calculated under multiple IoU thresholds (ranging from 0.5 to 0.95, with a step size of 0.05). From the Tab. \ref{tab:models_comparison}, it is evident that the metrics for two-stage and end-to-end methods are generally high, whereas the metrics for one-stage methods are comparatively lower. Our model as a one-stage method, achieve mAP@0.5 the indicator reached 48.8, mAP@0.5-0.95 the value reached 29.9, which was 1.1\% and 0.6\% higher in mAP 50, mAP 50:95, and mAP compared to the same one-stage YOLOv9 model. Furthermore, compared to the two-stage model, our model performed better on the mAP0.5 metric than the classic Faster R-CNN and Cascade R-CNN models. Compared to Cascade R-CNN, our model has improved by 0.7\% and 0.3\% respectively, demonstrating its outstanding performance in drone-based object detection tasks.

\begin{figure}[ht]
    \centering
    \includegraphics[width=1\linewidth]{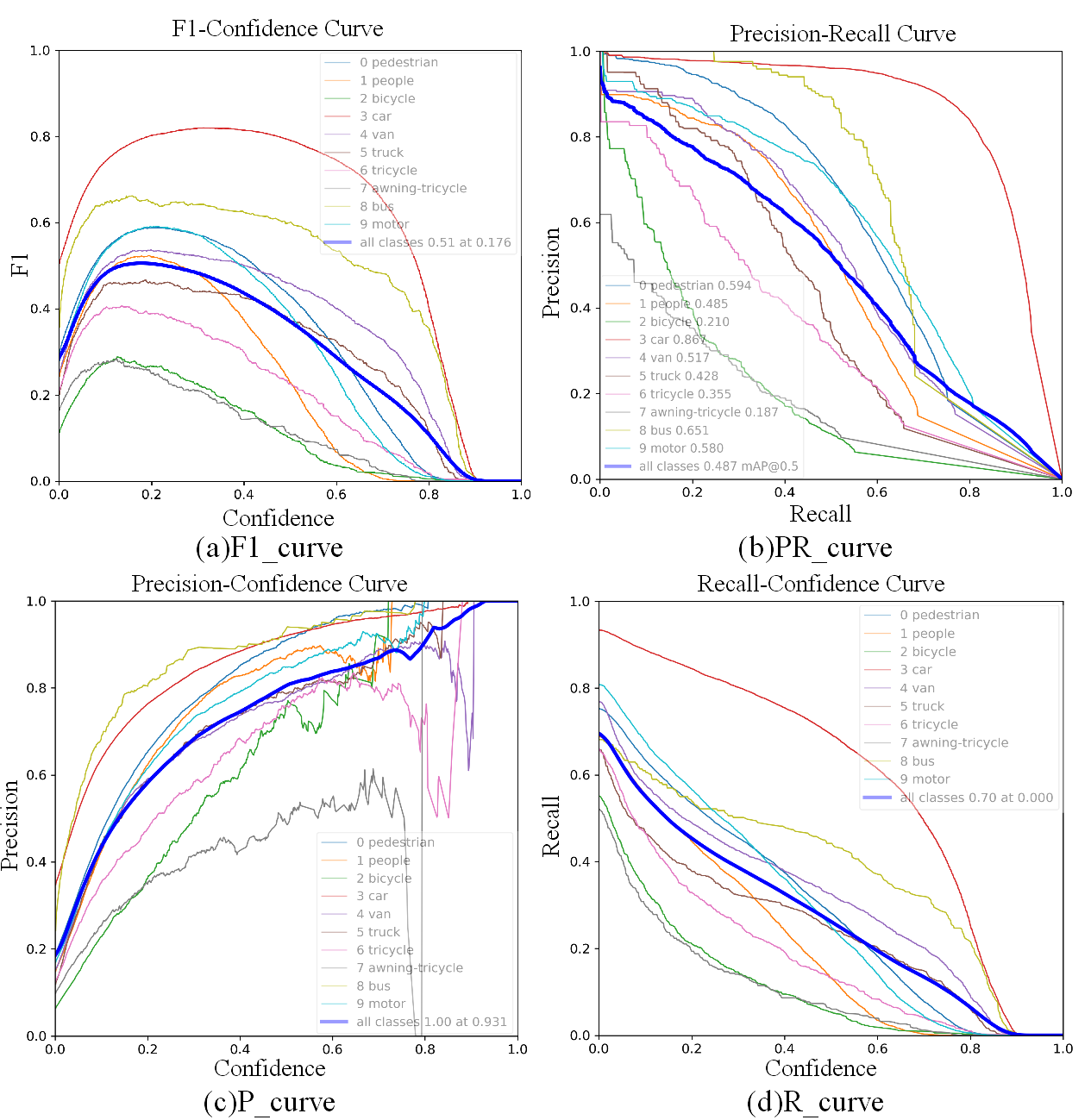}
     \caption{Visualization of evaluation parameters.(a)$F1$-curve.(b)$PR$-curve.(c)$P$-curve.(d)$R$-curve.}
    \label{fig:fig8}
\end{figure}

\begin{figure}[ht]
    \centering
    \includegraphics[width=1\linewidth]{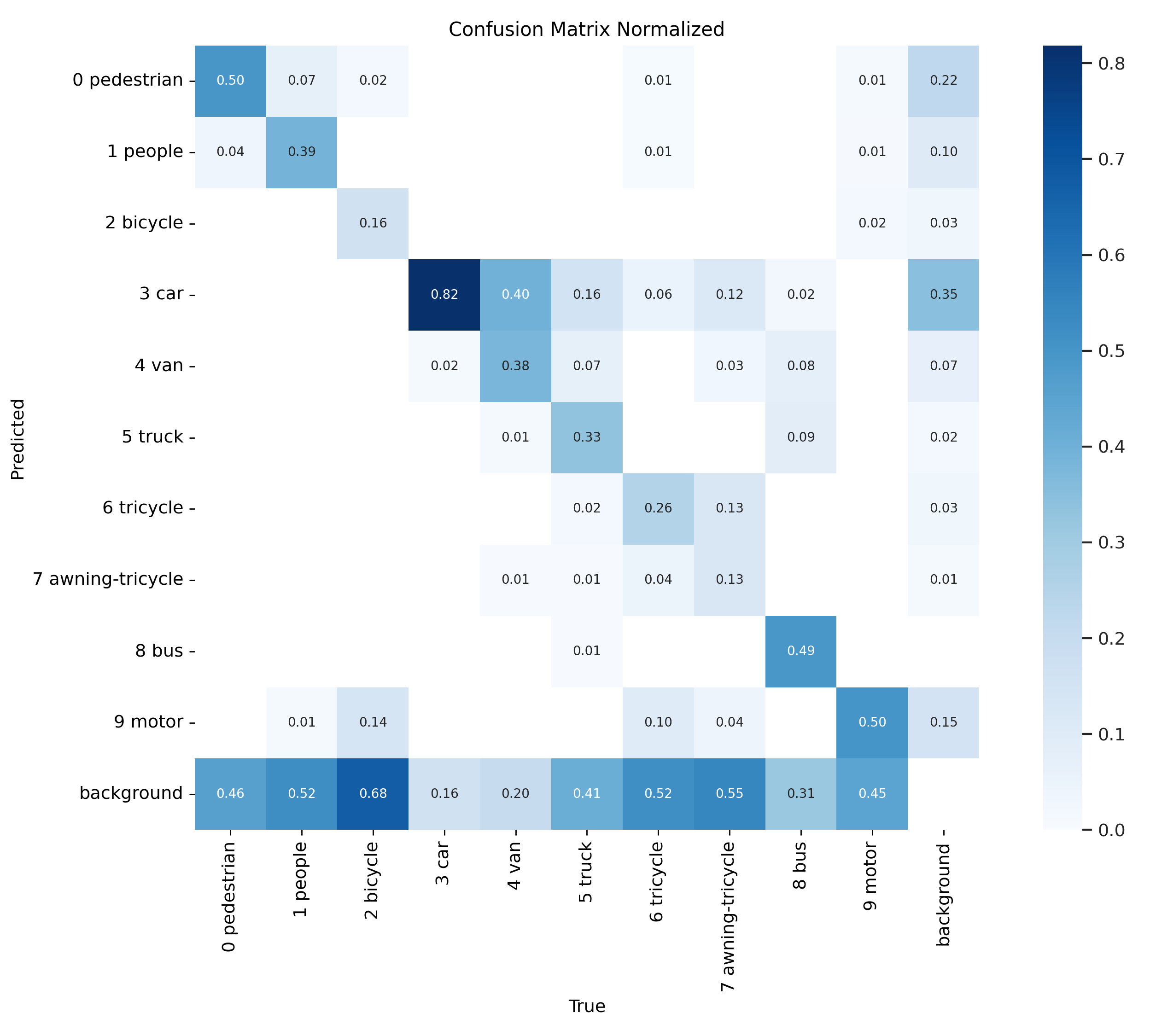}
    \caption{Confusion matrix diagram of LAM-YOLO model. The darker the color, the closer the correlation coefficient is to 1}
    \label{fig:fig9}
\end{figure}

\begin{figure*}
    \centering
    \includegraphics[width=1.0\linewidth]{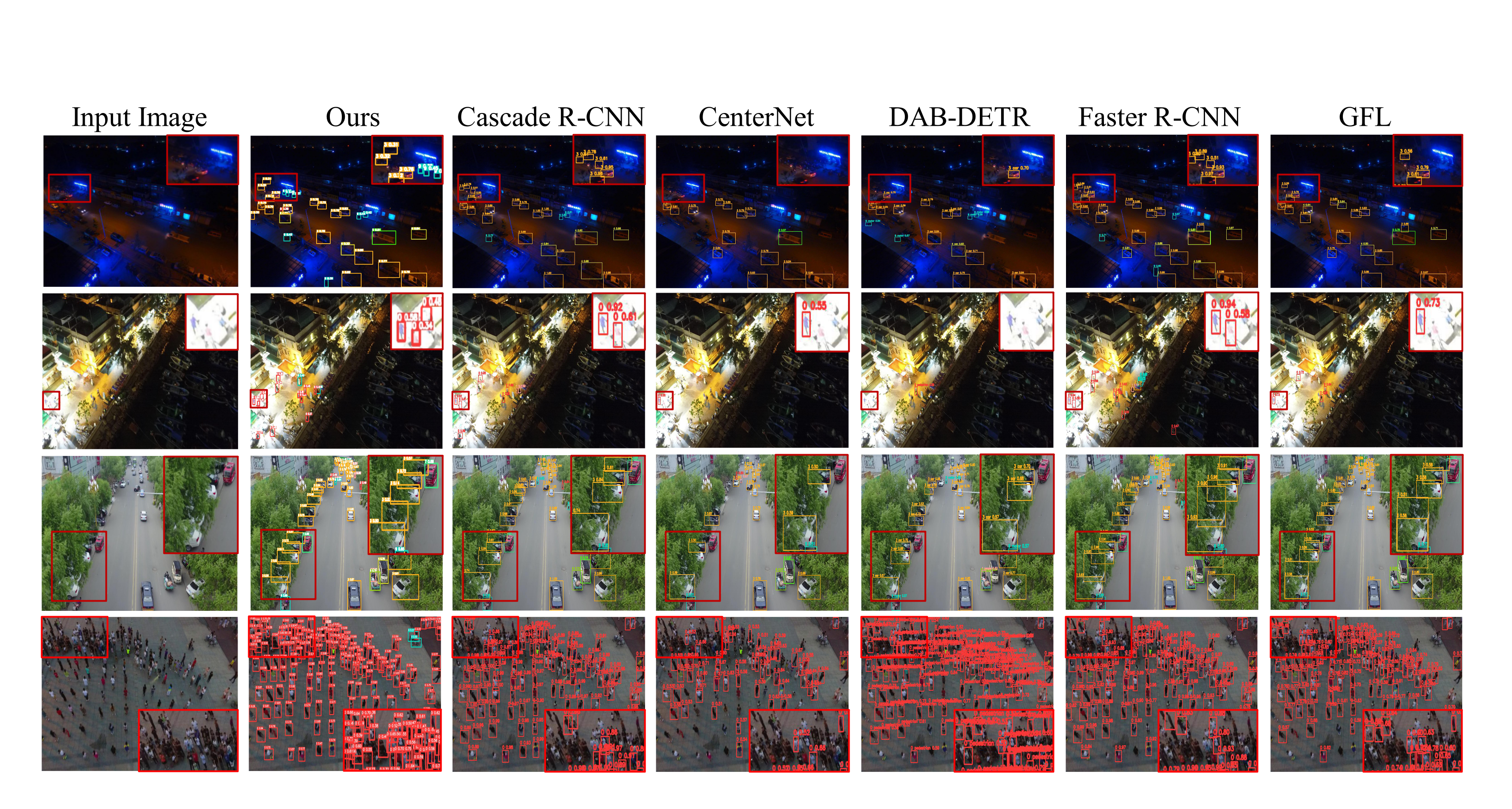}
    \caption{Improved visual results of model detection on data set VisDrone2019. The first line represents the recognition performance of different methods in low light environments, the second line represents the recognition performance of different methods in strong light environments, the third line represents the recognition performance of different methods in occluded environments, and the fourth line represents the recognition performance of different methods in dense environments. The red bounding box indicates a locally enlarged area.}
    \label{fig:fig10}
\end{figure*}

To further evaluate the detection performance of our model, we conducted comparative experiments focusing on three key aspects: evaluation metrics, confusion matrix, and model inference results. Fig. \ref{fig:fig8} presents the evaluation metrics of the improved model on the VisDrone2019 dataset. The graphs visualize key metrics, including the harmonized average, precision-recall (P-R) curve, accuracy, and recall, arranged from left to right and top to bottom. The analysis indicates that the LAM-YOLO achieves optimal detection accuracy while maintaining a high recall rate. The corresponding confusion matrix is depicted in Fig. \ref{fig:fig9}.

In summary, the results of the comparative experiments demonstrate that LAM-YOLO outperforms other models in detection performance. Our multi-scale feature fusion network enables five-scale detection and incorporates LAM, which provides distinct advantages in detecting drone-based small objects compared to the models evaluated in the aforementioned experiments.

\begin{figure*}[h]
    \centering
    \includegraphics[width=1.0\linewidth]{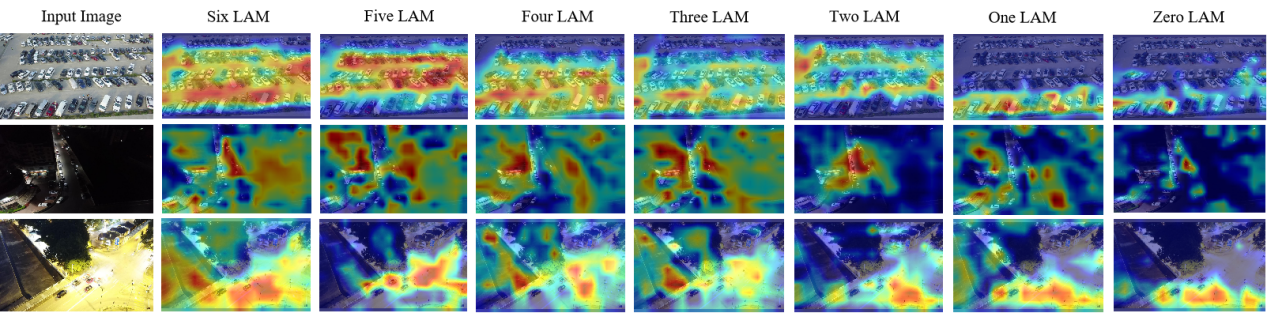}
    \caption{Improved visual results of model detection on data set VisDrone2019. The more prominent the heatmap, the stronger the feature extraction ability of the network for areas with dark and bright light.}
    \label{fig:figcam}
\end{figure*}

\begin{figure*}[h]
  \centering
    \vspace{-1\baselineskip}
  \subfloat[Loss-Epoch]{
    \includegraphics[width=.45\textwidth]{
      "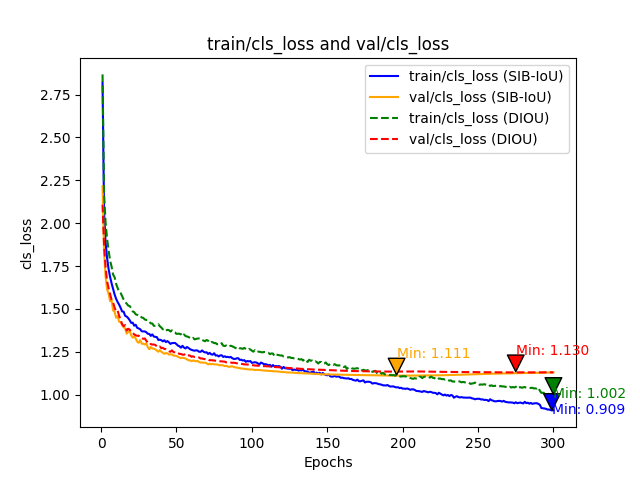"}
      \label{subfig:123}
      }
  \subfloat[MAP-Epoch]{
    \includegraphics[width=.45\textwidth]{
      "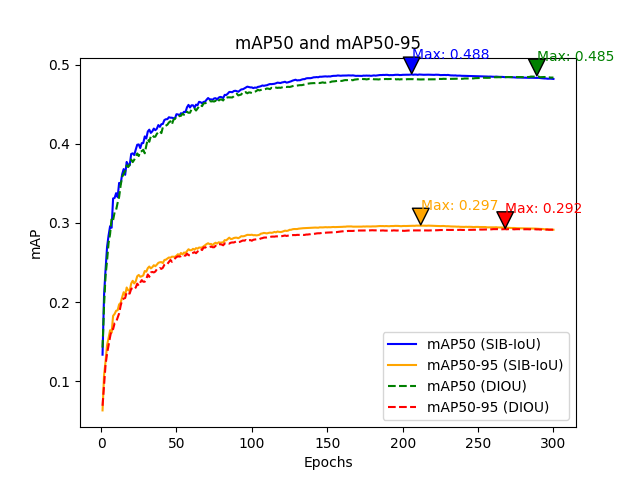"}
      \label{subfig:234}
      }
  \caption{Comparative Loss-Metric curve analysis of detection methods on (a) Loss-epochs and (b) MAP-epochs. Our SIB-IoU Loss demonstrates superior performance, as evidenced by its consistently higher training speed, highlighting it can enhance the detection performance.
  }
  \label{fig:siblossfun}
\end{figure*}

\begin{table*}[hb]
            \centering
            \footnotesize
            \caption{Comparison of detection accuracy between the improved model and YOLOv8s}
            \label{tab:improved model}
            \renewcommand{\arraystretch}{1.2} 
            \setlength{\tabcolsep}{1pt} 
            \resizebox{\linewidth}{!}{ 
                \begin{tabular*}{\dimexpr\textwidth+2\oddsidemargin\relax}{@{\extracolsep{\fill}} lcccccccccccc } 
                    \toprule
                    Class & \textbf{mAP $\uparrow$} & \textbf{Pedestrian $\uparrow$} & \textbf{People $\uparrow$} & \textbf{Bicycle $\uparrow$}& \textbf{Car $\uparrow$} & \textbf{Van $\uparrow$} & \textbf{Truck $\uparrow$} & \textbf{Tricycle $\uparrow$} & \textbf{Awning-Tricycle $\uparrow$} & \textbf{Bus $\uparrow$} & \textbf{Motor $\uparrow$} \\
                    \midrule
                    YOLOv8s & 41.7\% & 43.9\% & 35.2\% & 14.3\% & 80.0\% & 46.4\% & 42.2\% & 32.2\% & 18.2\% & 58.0\% & 46.8\%\\
                    Ours & 48.8\% & 59.6\% & 48.7\% & 21.1\% & 86.8\% & 51.8\% & 42.8\% & 35.5\% & 18.7\% & 65.1\% & 58.3\% \\ \rowcolor[rgb]{0.9,0.9,0.9}
                    improve & 7.1\% & 15.7\% & 13.5\% & 6.8\% & 6.8\% & 5.4\% & 0.6\% & 3.3\% & 0.5\% & 7.1\% & 11.5\% \\
                    \bottomrule
                \end{tabular*}
            }
\end{table*}

\subsubsection{\textbf{Qualitative comparisons and analysis}}To evaluate the effectiveness of LAM-YOLO, we conducted a comprehensive comparison across low-light scenes, strong light environments, dense crowd scenes, and complex occluded road scenes at night. In these different environments, our model demonstrated superior detection performance, significantly improving its ability to accurately predict target categories, recognize vehicles and other objects accurately. The effectiveness of the improved detection model is further confirmed by the visual results presented in Fig. \ref{fig:fig10}. However, while Cascade R-CNN can detect most targets, it still encounters issues with missed detections and inaccurate bounding boxes in certain areas. The CenterNet demonstrates suboptimal performance in handling occlusions and dense scenes, resulting in increased missed detections and inadequate performance in low-light conditions. DAB-DETR exhibits inadequate performance in these scenarios, characterized by false detections and missed detections, particularly in strong light environments. Faster R-CNN demonstrates relatively stable performance in occluded and nighttime scenes but exhibits deficiencies in other scenarios. While GFL can recognize some targets, its detection accuracy in dense areas is relatively low, indicating shortcomings in its sensitivity to small target detection.

Our proposed model demonstrates superior performance across various complex environments. The improved network effectively captures global interdependence, reduces the missed detection rate for occluded and densely packed objects, and significantly enhances overall detection performance. Its detection accuracy notably surpasses that of other mainstream object detection algorithms. This indicates that our model has achieved substantial improvements in feature extraction, target localization, and occlusion processing capabilities, thereby better addressing various challenges in practical applications.

\subsection{Ablation Study} \label{subsec:ablation}
\subsubsection{\textbf{Ablation of different Model Components}} \label{subsec:lamablation} We validate the effectiveness of each proposed improvement strategy and conducted ablation experiments on the model. The results are shown in Tab. \ref{tab:Ablation experiment}, each improvement strategy has correspondingly improved detection performance. The first improvement is the addition of two detection heads, sized 160×160 and 320×320, which are particularly effective for detecting small targets, a critical need given the prevalence of small objects in the VisDrone2019 dataset. The P2 detection head improved accuracy by 3.4\%, and the P1 detection head by 1.6\%. Subsequently, the Lighting-occlusion attention module(LAM) mixed attention module was incrementally integrated into each output layer, leading to further performance gains. The Involution module also contributed to a relative accuracy improvement, particularly in reducing the omission rate of small targets. Overall, the enhanced model achieved a 7.1\% increase in average detection accuracy, with significant improvements across most detection metrics.

To demonstrate the effectiveness of the improved model in enhancing detection performance, we conducted a comparative experiment between the improved model and YOLOv8s. As indicated by the comparison results in Tab. \ref{tab:improved model}, the AP values for each category and the mAP@0.5 values across all categories for both models. Moreover, the mAP of the improved model increased by 7.1\%. The AP values for all categories exhibited varying degrees of improvement, with the Pedestrian, People, and Motor categories experiencing increases exceeding 10\%. These results indicate that the improved model effectively enhances detection accuracy.

Moreover, Fig. \ref{fig:figcam} provides a more intuitive display of the impact of different numbers of LAM on feature extraction in the input image through Grad-CAM. As the number of LAM layers gradually decreases from "Six LAM" to "Zero LAM", the highlighted areas in the heatmap become weaker and unevenly distributed, indicating that the model's attention to important features is decreasing. In various scenarios, as LAM decreases, the attention area to the target also shrinks, and the focus on distant targets gradually diminishes.

\begin{table}[h]
    \centering
    \footnotesize
    \caption{Performance comparison of YOLOv8 models of different sizes on the VisDrone2019 dataset}
    \label{tab:yolov8}
    \begin{tabular}{cccc}
        \hline
        \textbf{Model} & \textbf{mAP0.5/\%} & \textbf{Parameter/$10^6$} & \textbf{FLOPS(B)} \\
        \hline
        YOLOv8n & 35.1 & 3.01 & 8.7 \\
        YOLOv8s & 41.7 & 11.14 & 28.7 \\
        YOLOv8m & 45.1 & 25.86 & 79.1 \\
        YOLOv8l & 46.9 & 43.63 & 165.4 \\
        \hline
    \end{tabular}
\end{table}

\subsubsection{\textbf{Ablation of backbone}} 

Building upon its predecessor, YOLOv8 introduces new features that enhance its suitability for a wide range of object detection tasks across various applications. The official release includes four distinct YOLOv8 models: YOLOv8n, YOLOv8s, YOLOv8m, and YOLOv8l, each differing in size and complexity. The parameters of these models are detailed in Tab. \ref{tab:yolov8}. Given that the model is intended for deployment on drones, it is essential to achieve a careful balance between accuracy and complexity. Among the YOLOv8 models, YOLOv8s offers the optimal compromise by minimizing complexity while preserving accuracy. Consequently, our model is developed based on the YOLOv8s architecture.

To verify the effectiveness of the improved model in enhancing detection performance, a comparative experiment was conducted between the improved model and YOLOv8. As shown in the comparison results in Tab. \ref{tab:improved model}, the mAP of the improved model increased by 7.1\%. All categories exhibited varying degrees of improvement in AP values, with the pedestrian, human, and motorcycle categories showing increases of over 10\% in AP values. The experimental results demonstrate that the improved model effectively enhances detection accuracy.

\begin{table}[h]
    \centering
        \footnotesize
    \caption{Comparison of detection results of different loss functions introduced by the improved model}
    \label{tab:loss}
    \begin{tabular}{cccc}
        \hline
        \textbf{Metrics} & \textbf{mAP0.5/\%} & \textbf{mAP0.75/\%} & \textbf{mAP0.5:0.95/\%} \\
        \hline
        WIoUv3 \cite{Wiou}& 48.2 & 30.4 & 29.4 \\
        CIoU \cite{Zheng2020}& 48.1 & 30.0 & 29.3 \\
        DIoU \cite{Zheng2020}& 48.0 & 29.9 & 29.4 \\
        GIoU \cite{GIOU}& 48.1 & 30.5 & 29.6 \\
        EIoU \cite{EIOU}& 47.6 & 30.1 & 29.3 \\
        SIoU \cite{Gevorgyan2022}& 48.5 & 30.4 & 29.6 \\ \rowcolor[rgb]{0.9,0.9,0.9}
        \textbf{SIB-IoU} & \textbf{48.8} & \textbf{30.7} & \textbf{29.9} \\
        \hline
    \end{tabular}
\end{table}

\subsubsection{\textbf{Ablation of SIB-IoU Loss Function}} \label{subsec:lossablation}
To validate the benefits of incorporating SIB-IoU, we conducted comparative experiments using SIB-IoU and other mainstream loss functions on the improved model, while maintaining consistent training conditions. The results are presented in Tab. \ref{tab:loss}. The experiments demonstrate that the model achieves optimal detection performance when SIB-IoU is used as the bounding box regression loss. Furthermore, the mAP50 of the model with SIB-IoU was 0.7\% higher than with CIoU, highlighting the effectiveness of SIB-IoU.

In a separate set of experiments, we verified the benefits of introducing SIB-IoU using the improved model with SIB-IoU and other mainstream loss functions, as illustrated in Fig. \ref{fig:siblossfun}, indicate that the model's detection performance is optimal when SIB-IoU serves as the bounding box regression loss. Additionally, the mAP50 of the model using SIB-IoU is 0.8\% higher than that of the model using DIoU, further emphasizing the effectiveness of SIB-IoU. The graph illustrates that the model utilizing SIB-IoU achieves a faster rate of loss reduction and a more rapid increase in mAP values.

\section{Conclusion} \label{sec:conclusion}

This paper presents LAM-YOLO, a target detection model specifically designed for drone-based aerial scenes. First, the LAM attention mechanism is integrated into the backbone network and neck, emphasizing key information within the feature maps to assist the model in accurately identifying and locating targets. Additionally, Involution modules are incorporated into the neck to enhance the interaction among features at different levels. Furthermore, we adopt the SIB-IoU loss function to generate auxiliary bounding boxes at various scales, effectively improving the model's ability to detect targets of different sizes. Finally, two additional auxiliary detection heads are introduced to further expand the model's detection capabilities. This approach integrates both shallow and deep features, significantly reducing the miss rate for small objects and greatly enhancing detection accuracy under high and low lighting condition. Experimental results demonstrate that the average detection precision of the LAM-YOLO model improves significance compared to the baseline model. Overall, Future research should focus on optimizing detection accuracy while managing calculate resource consumption effectively.


\bibliographystyle{IEEEtran}
\bibliography{./reference.bib}

\end{document}